\documentclass[conference]{IEEEtran}
\IEEEoverridecommandlockouts
\usepackage{cite}
\usepackage{amsmath,amssymb,amsfonts}
\usepackage{algorithmic}
\usepackage{graphicx}
\usepackage{textcomp}
\usepackage{adjustbox}
\usepackage{float}
\usepackage{xcolor}
\def\BibTeX{{\rm B\kern-.05em{\sc i\kern-.025em b}\kern-.08em
T\kern-.1667em\lower.7ex\hbox{E}\kern-.125emX}}

\usepackage{hyperref}
\hypersetup{
    colorlinks=true,
    linkcolor=black,
    filecolor=black,      
    urlcolor=black,
    citecolor=black
    }
    
\usepackage{microtype}
\usepackage[none]{hyphenat}

\usepackage{colortbl}
\newcolumntype{M}[1]{>{\centering\let\newline\\\arraybackslash\hspace{0pt}}m{#1}}

\usepackage{enumitem}

\makeatletter
\newcommand\BeraMonottfamily{%
  \def\fvm@Scale{0.85}
  \fontfamily{cmtt}\selectfont
}
\makeatother

\usepackage{listings}
\renewcommand{\baselinestretch}{1.05}

\begin{document}

\title{Utilizing deep learning for automated tuning of database management systems}

\author{
\adjustbox{max width=\textwidth}{
\begin{tabular}{ccc}
 Karthick Gunasekaran* &&  Kajal Tiwari \\ 
\textit{Manning College of Information and Computer Sciences} && \textit{Manning College of Information and Computer Sciences}  \\ 
\textit{University of Massachusetts Amherst} && \textit{University of Massachusetts Amherst}
\end{tabular}}
\\[5mm]
\adjustbox{max width=\textwidth}{
\begin{tabular}{c}
 Rachana Acharya \\ 
\textit{Manning College of Information and Computer Sciences} \\ 
\textit{University of Massachusetts Amherst}
\end{tabular}}
}

\maketitle

\begin{abstract}
Managing the configurations of a database system poses significant challenges due to the multitude of configuration knobs that impact various system aspects. The lack of standardization, independence, and universality among these knobs further complicates the task of determining the optimal settings. To address this issue, an automated solution leveraging supervised and unsupervised machine learning techniques was developed. This solution aims to identify influential knobs, analyze previously unseen workloads, and provide recommendations for knob settings. The effectiveness of this approach is demonstrated through the evaluation of a new tool called OtterTune \cite{Dana_Van} on three different database management systems (DBMSs). The results indicate that OtterTune's recommendations are comparable to or even surpass the configurations generated by existing tools or human experts. In this study, we build upon the automated technique introduced in the original OtterTune paper, utilizing previously collected training data to optimize new DBMS deployments. By employing supervised and unsupervised machine learning methods, we focus on improving latency prediction. Our approach expands upon the methods proposed in the original paper by incorporating GMM clustering to streamline metrics selection and combining ensemble models (such as RandomForest) with non-linear models (like neural networks) for more accurate prediction modeling.
\end{abstract}


\section{Introduction}
In both business and scientific fields, the ability to gather, process, and analyze large amounts of data is crucial for gaining new knowledge \cite{Dworin,Laney}. To accomplish this, data-intensive applications rely heavily on database management systems (DBMSs), which are responsible for collecting, storing, and retrieving data. However, achieving good performance in DBMSs can be challenging, as they are complex systems with many configuration options that can affect their runtime behavior. Although these options can be adjusted by a database administrator, the correct configuration depends on various factors and is difficult to reason about.

 DBMSs are intricate systems with numerous adjustable options that regulate almost all aspects of their runtime operation \cite{24_cite}. These configurable options empower database administrators (DBAs) to manage various aspects of the DBMS's runtime behavior, such as allocating memory for data caching versus the transaction log buffer. Modern DBMSs are known for having a multitude of configuration knobs, as highlighted in \cite{22_cite, 47_cite, 36_cite }. The challenge in achieving optimal performance in DBMSs stems from the fact that their performance and scalability are heavily influenced by their configurations\cite{24_cite}. Furthermore, the default configurations of these knobs are notoriously suboptimal, compounding the issue. As an illustration, the default MySQL configuration in 2016 assumes that the system is deployed on a machine with only 160 MB of RAM \cite{1_cite}.

Automatic tuning tools have been developed to optimize DBMS performance, but most of them have limitations, such as being designed for specific DBMSs or requiring manual steps. In OtterTuner paper\cite{Dana_Van}, the authors propose a new approach that utilizes machine learning models to reuse training data from previous tuning sessions. This technique selects the most important configuration options, maps new workloads to known ones, and recommends optimal settings based on a target objective (e.g., latency or throughput). By reusing past experience, the proposed method reduces the time and resources needed to optimize DBMS performance for new applications.

The approach was implemented in a tuning tool called OtterTune, using Google TensorFlow and Python's scikit-learn libraries. Experiments were conducted on two OLTP DBMSs (MySQL and Postgres) and one OLAP DBMS (Vector), and the results show that OtterTune can achieve significant improvements in latency compared to default settings or other tuning advisors. Furthermore, the tool generates configurations within an hour that are comparable to those created by expert DBAs.

The main objective of our work is to extend the OtterTuner and propose novel machine learning (ML) models from data collected from previous tunings and use the models primarily for (1) pruning the redundant metrics, (2) mapping unseen database workloads to previous workloads from which we can transfer experience, and (3) improving latency prediction through workload mapping.
We have achieved this with the following steps: 

\begin{itemize}
\item \textit{Pruning Redundant Metrics:} Since we have hundreds of internal and external DBMS-specific metrics, we limit our search space to a small set, which impacts latency significantly. This step speeds up the entire process and ensures that the model will fit in memory.
\item  \textit{Workload Mapping:} Workload mapping is the process of matching the target DBMS’s workload with the most similar workload in its repository based on performance measurements. This is done by finding the nearest neighbor with the help of Euclidean distance and calculating the "score" for each workload by taking
the average of these distances over all metrics. Reusing experience reduces time and resources and also helps the model make an educated decision in the case of fewer observations in the current workload.
\item  \textit{Latency Prediction:} The ultimate goal is to improve latency prediction through workload mapping. Prediction modeling will be done through the regression model-Guassian Process Regression), but we'll also be experimenting with neural networks.
\end{itemize}

\section{Literature Review}
In the past 50 years, researchers have explored methods to optimize DBMSs automatically [\cite{paper_12}, \cite{paper_18}, \cite{paper_16}, \cite{paper_74}]. This body of work includes theoretical and applied research [\cite{paper_14}, \cite{paper_24}, \cite{paper_82}, \cite{paper_75}]. Prior to the 2010s, automated DBMS tuning primarily relied on heuristic or cost-based algorithms. Heuristic algorithms, such as IBM's DB2 Performance Wizard, used hard-coded rules to recommend actions based on application characteristics [\cite{paper_32}, \cite{paper_42}]. However, these rules may not accurately reflect the actual workload and environment. Some vendors, like IBM and Oracle, introduced self-tuning mechanisms based on rules to allocate memory and identify misconfiguration bottlenecks [\cite{paper_67}, \cite{paper_70}, \cite{paper_19}, \cite{paper_21}, \cite{paper_40}]. Cost-based algorithms, on the other hand, programmatically searched for configuration improvements guided by a cost model derived from workload traces [\cite{paper_17}, \cite{paper_58}, \cite{paper_82}, \cite{paper_54}, \cite{paper_77}, \cite{paper_13}]. To address the limitations of cost-based algorithms, using accurate cost model estimations from the DBMS's internal components proved beneficial, as demonstrated by Microsoft's AutoAdmin utilizing SQL Server's built-in cost models for index utility estimation [\cite{paper_15}, \cite{paper_65}]. However, such reliance on the DBMS cost model is not suitable for knob configuration tuning algorithms. Knobs, which affect DBMS behavior, are not easily reflected in the query planner's cost model due to their varying impact based on the workload [\cite{paper_65}]. As a result, major vendors' proprietary knob configuration tools mainly employ static heuristics and vary in the level of automation [\cite{paper_21}, \cite{paper_40}, \cite{paper_53}, \cite{paper_2}, \cite{paper_7}]. Both heuristic and cost-based tuning methods have the drawback of focusing on individual DBMS instances and lacking the utilization of information from previous tuning sessions. Heuristic-based approaches make assumptions about workload and environment, which may not accurately represent the real-world scenario. Despite these optimizations, solving database tuning problems efficiently remains challenging due to their NP-Complete nature. Machine learning (ML)-based approaches can potentially offer faster approximations for optimization problems, addressing this challenge.

\section{Approach}
The Overview of the entire system can be seen from the \autoref{fig:archi} below. The process consisted of two important steps. In the first step, important metrics were pruned, and in the next step, automated tuning was carried out through workload mapping.   
\begin{figure}[H]
\centering
\includegraphics[width=\linewidth]{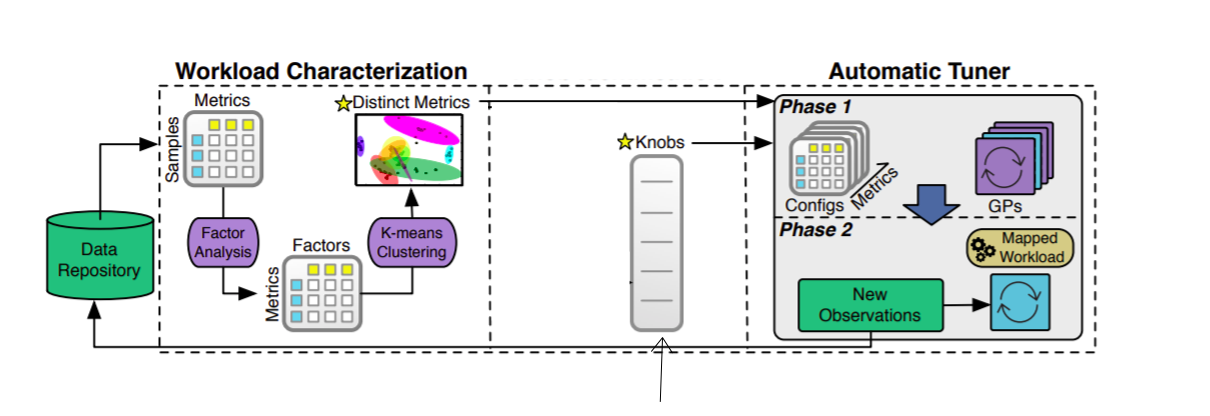}
\caption{Overview of the architecture \cite{Dana_Van}}
\label{fig:archi}
\end{figure}

\subsection{Metrics pruning}

In this stage, pruning of redundant metrics was carried out. In order to capture the variability of system performance and differentiate different workloads, a small number of metrics with high variability are considered.  The system needs to capture a few metrics such that all the distinguishing characteristics of workloads are captured while the overall runtime of the machine learning algorithms reduces. The pruning step is very important because it helps improve the entire performance and speed of the system. Pruning is carried out in two steps. In the first step, factor analysis is carried out, followed by K-means clustering.

\subsubsection{Data Preprocessing}

We had to preprocess the data present in the offline workload and the online workloads B and C. Data pre-processing was done in the following steps:

\begin{itemize}
\item \textit{Duplicate Column Removal:} We dropped the columns or metrics with only a constant value across all workloads.
\item \textit{Conversion of boolean knobs:} We also did some other preprocessing steps, such as converting all boolean knob values to integers (0 and 1).
\item \textit{Division of files by workloads:}  We segregated all files on the workload ID; hence, for offline workloads, we had 58 workload files, for online workload C, we had 100 workload files; and for online workload B, we divided each workload into 5 rows for workload mapping and 1 row from each workload for validation.
\end{itemize}

\subsubsection{Factor Analysis}

In this step, the strongly correlated metrics are removed as they're redundant. A dimensionality reduction technique called factor analysis \cite{sklearn_FA} is employed to transform the high dimensional data into low-dimensional data.

Each factor is a linear combination of the original variables; the factor coefficients are similar to and can be interpreted in the same way as the coefficients in a linear regression. Furthermore, each factor has a unit variance and is uncorrelated with all other factors. 

The FA algorithm takes as input a matrix $X$ whose rows correspond to metrics and whose columns correspond to knob configurations that we have tried. The entry $X_{ij}$ is the value of metric $i$
on configuration $j$. FA gives us a smaller matrix $U$: the rows of
$U$ corresponds to metrics, while the columns correspond to factors,
and the entry $U_{ij}$ is the coefficient of metric $i$ in factor $j$. We can
scatter-plot the metrics using elements of the $i^{th}$ row of $U$ as coordinates for metric $i$. Metrics $i$ and $j$ will be close together if they
have similar coefficients in $U$ - that is, if they tend to correlate
strongly in $X$. Removing redundant metrics now means removing
metrics that are too close to one another in our scatterplot.

\begin{figure}[htbp]
\centering
\includegraphics[width=\linewidth]{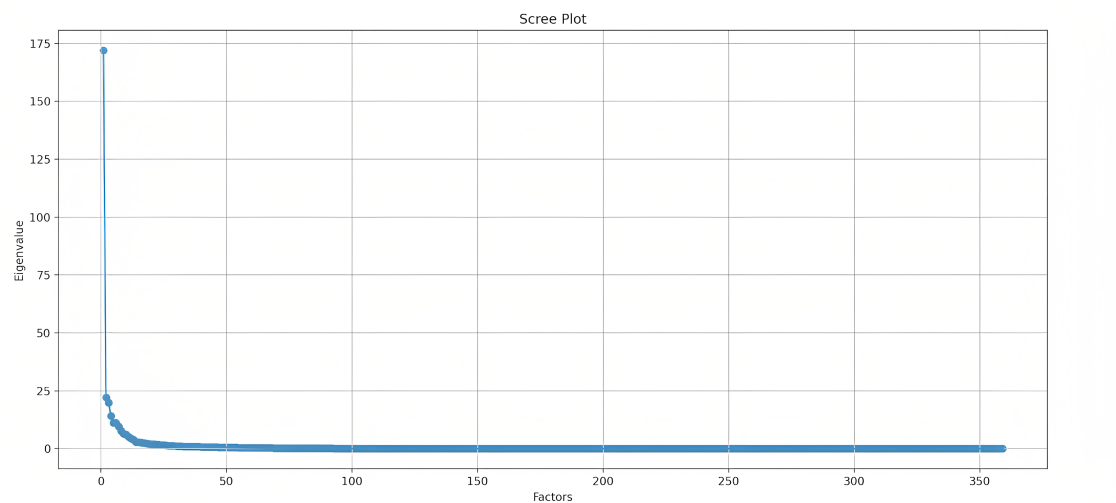}
\caption{Eigen values of every factor}
\label{fig:scree_plot}
\end{figure}

The data from offline workloads, which consist of 58 workloads, is used for this. Here, all the data related to the knobs is removed, and this truncated data is provided as input to the factor analysis module. Having a look at the eigen values of all the factors generated, only the initial 30 factors with eigen values greater than 1 are observed to be significant for the DBMS metric data. Therefore, only the most significant factors are taken into account.
\autoref{fig:scree_plot} shows a eigen values where the factors are plotted against their respective eigen values.

\subsubsection{K-means Clustering}

K-means clustering is used to identify and segregate metrics into meaningful groups. It works by clustering data to separate it into n groups of equal variance. The algorithm works towards minimizing the distance between the data points within the cluster. The input to the K-means model consists of a matrix comprising our 30 factors from factor analysis. A metric closest to each of the cluster centers can be picked up. Sklearn's implementation of the K-means algorithm \cite{sklearn_kmeans} was used.

K-means doesn't provide the optimal number of clusters. Initially, K-means clustering is carried out with different cluster numbers from 2 to 15. Then the optimal number of clusters is identified. Instead of doing manual interpretations of the clusters, silhouette analysis was carried out to study the separation distance between the clusters and  identify the optimal clusters. Silhouette values are determined by measuring the closeness of the points in a cluster to its neighboring clusters. The values range from -1 to +1. The value closer to 1 indicates the decision boundary is too far from the neighboring clusters, while values close to -1 indicate the clusters are too close to each other. So, silhouette scores for KMeans models with different cluster configurations ranging from 2 to 15 were calculated. The highest silhouette score near +1 indicates the best cluster configuration and its selection. 

Since the optimal number of clusters was identified as 8, the metrics are segregated into 8 clusters, and the centroids are computed for the same. One metric from each cluster that is closest to the centroid was selected. This metric will represent the entire cluster. Once we find our target metrics, we remove all the other redundant metrics from the data and proceed to the next step.

List of pruned metrics by this process include,

{\scriptsize \begin{itemize}[itemsep=-9pt]

\item 
\begin{lstlisting} 
driver.jvm.pools.Code-Cache.committed.avg_inc
\end{lstlisting}

\item 
\begin{lstlisting}
driver.jvm.pools.Code-Cache.committed.avg_period
\end{lstlisting}	

\item 
\begin{lstlisting}
driver.BlockManager.memory.maxMem_MB.avg
\end{lstlisting}

\item 
\begin{lstlisting}
driver.BlockManager.memory.onHeapMemUsed_MB.avg
\end{lstlisting}

\item 
\begin{lstlisting}
driver.LiveListenerBus.queue.executorMgmt.size.avg
\end{lstlisting}

\item 
\begin{lstlisting}
executor.jvm.pools.PS-Old-Gen.cmt.avg
\end{lstlisting}

\item 
\begin{lstlisting}
worker_1.Disk_transfers_per_second.dm-0
\end{lstlisting}

\item 
\begin{lstlisting}
worker_1.Paging_and_Virtual_Memory.pgpg
\end{lstlisting}

\end{itemize}}

\subsection{Automated Tuning}

\subsubsection{Preprocessing}

All the pruned metrics other than the "latency" metric and the  knobs were normalized by removing the mean and scaling it to unit variance. Zero mean scaling is a very common practice before using the data in machine learning algorithms. This step is also essential to avoid overflowing values while calculating the mean squared error. 

\subsubsection{Gaussian Process Regression}

Gaussian process regression, or GPR \cite{sklearn_GPR} a probabilistic-based supervised learning approach, was used to train the model. GPR requires prior specification. The prior in this case is the offline workloads, which were seen before and were fitted initially. The covariance of the prior is specified by the kernel chosen. The log marginal likelihood is maximized to optimize the hyperparameters during the fitting. Two different kernels were tried, and the one that performed best was chosen. Radial basis function (RBF) performed better compared with others. The scaled knobs and metrics were used to fit the model. Sklearn's implementation of the Gaussian process regression \cite{sklearn_GPR} as well as Ottertuner implementation of the GPR were also used.
\subsubsection{Workload Mapping}

For workload mapping, we match the current target workload to offline workloads B and C. We calculate a "score" for each workload by averaging the distances obtained using Euclidean distance across all metrics.

The step-by-step process followed for workload mapping is:

\begin{itemize}
\item  Calculate the Euclidean distance between the metrics vector for the target workload (workload B in the first iteration) and the corresponding vector for
each workload in the offline workload. 
This computation is done for each metric.
\item  Compute a 'score' for each workload i by taking the average of these distances over all metrics. The algorithm then chooses the workload with the lowest score as the one that is most similar to the target workload.
\item Augment the target workload to the nearest source workload; in case of a conflict of knob configurations between the nearest source and the target workload, keep the configurations of the target workload.
\item Train the models on 'Augmented workloads' and predict the latency on the validation set created from online workload B.
\item Repeat the process with the source workload as 'augmented workload' and the target workload as 'Workload C' and create the final augmented workload for training the models and the final prediction on test.csv.
\end{itemize}

\subsubsection{Latency Prediction}

Latency prediction is carried out in two stages. In the first stage, the 6th row of all the workloads in the online\_workloadB is used to predict the latency while trained on offline workloads appended with workload-mapped data from online\_workloadB.The latency values are then estimated. In the next stage, the test.csv workloads are predicted with the model trained on data appended from both workloads B and C. This process adds more data for the GPR model to make accurate predictions.

\section{Extensions}

\subsection{GMM Clustering}

The original OtterTune paper \cite{Dana_Van} makes use of Kmeans for clustering metrics. But, there are certain disadvantages to using K-means. It assumes that the clusters always have a spherical shape. Moreover, it takes only the mean of the clusters into account and not their variance. Also, each cluster has roughly equal numbers of observations. Additionally, it is very sensitive to outliers. To overcome these inadequacies, we replace the Kmeans clustering with GMM clustering \cite{sklearn_GMM}.

GMM clustering overcomes these disadvantages by assuming each cluster as a different gaussian distribution and grouping data points belonging to a single distribution together. It therefore generates better clusters as it takes into account variance along with the mean of the clusters. One can think of mixture models as generalizing k-means clustering to incorporate information about the covariance structure of the data as well as the centers of the latent Gaussians. The GaussianMixture model used by us from sklearn \cite{sklearn_GMM} implements the expectation-maximization (EM) algorithm for fitting mixture-of-Gaussian models. Expectation-maximization is a well-founded statistical algorithm to get around this problem through an iterative process. The first one assumes random components (randomly centered on data points, learned from k-means, or even just normally distributed around the origin) and computes for each point a probability of being generated by each component of the model. Then, one tweaks the parameters to maximize the likelihood of the data given those assignments. Repeating this process is guaranteed to always converge to a local optimum. The GaussianMixture module comes with different options to constrain the covariance of the difference classes estimated: spherical, diagonal, tied, or full covariance. The full covariance option is made available by us, where each component has its own general covariance matrix, so the clusters may independently adopt any position and shape. The optimal number of clusters was found by making use of the Silhoutte and BIC scores for GMM. 

\subsection{Random Forest}

The random forest algorithm was tried instead of Gaussian process regression. One of the advantages of random forest over Gaussian process regression is that it performs very well with high-dimensional data. Also, the random forest is based on the bagging algorithm and uses the emsemble technique, where it creates many trees on a subset of the data and combines all the output. The overfitting is reduced through this technique, while GPR provides no such mechanism as the prior is fitted initially. Another reason to try random forest is to make use of a frequentist-based approach rather than a probabilistic approach such as GPR since frequentist based approaches perform better when there is more data. Also, due to their complexity, GPR requires much more time to train than random forests. Even though in this work comparatively less data is being used, when there is more data, random forest can be the right technique and might be preferable. The random forest algorithm \cite{sklearn_RF} was implemented with depths up to 50 levels and 200 estimators.

\subsection{Neural Network}

One major disadvantage with GPR used in the OtterTune paper\cite{Dana_Van} is that it doesn't perform well with higher-dimensional data. So the number of metrics has to be pruned and used in OtterTuner. To overcome this disadvantage, neural networks were used as an alternative to see the performance in higher dimensions. An artificial neural network has the ability to learn and model non-linear and complex relationships that contain trainable parameters. Moreover, an artificial neural network’s outputs aren't entirely limited by the inputs and results given to it initially by an expert system. Also, artificial neural networks have the ability to generalize their inputs. After learning from the initial inputs and their relationships, it can infer unseen relationships on unseen data as well, thus making the model generalize and predict on unseen data.

A complex 5-layer neural network model is created with a fully connected hidden layer. The Keras library is used for this purpose. No activation function is used for the output layer, as we are interested in predicting numerical values directly without transformation. The efficient ADAM optimization algorithm is used, and the MAPE loss function is optimized. This metric was specifically chosen because it is the one that will be used to compare all of our models.

\section{Experiments}

\subsection{Evaluation}

In K-means clustering, the silhoutte analysis was carried out to evaluate the optimal number of clusters in the data.

In the workload mapping stage, both mean squared error and mean average percentage error were tried for evaluation purposes. From the latency prediction results, it was observed that the MSE evaluation criteria in workload mapping performed better. Latency prediction evaluation was carried out using MAPE.

\subsection{Scaling and Hyperparameter tuning}

Experiments were carried out with and without scaling. Two different scaling variations were performed and analyzed. In the first variation, only the inputs for workload mapping and GPR models were tried. In the second variation, all the data except the latency column was scaled. In general, scaling all the data except the latency column with a zero mean and unit variance tends to improve the performance of the system. When unscaled data was used, it was observed that the predictions were far apart from the ground truth in general. 

Hyperparameter tuning was carried out for GPR. In the case of GPR, the noise levels were controlled by setting the alpha parameter to lower values. From the \autoref{tab:gpralpha} it could be seen that the model performed well as the alpha values are low. Random forest trees were built for different depths, and different estimators were also evaluated. The ones with the best evaluation scores were chosen as hyperparameters.

\begin{table}[htbp]
\caption{GPR hyperparameter tuning}
\label{tab:gpralpha}
\centering
\adjustbox{max width=\linewidth}{\fontsize{8}{11}\selectfont \it
\begin{tabular}{M{4cm}M{4cm}} \hline \hline 
\textbf{Alpha} & \textbf{MAPE} \\ \hline 
1e+8 &  168.63 \\  
1e+7 &  142.01 \\  
1e+5 &  123.67 \\  
1e+3 &  101.29 \\  
1e+1 & 98.45 \\  
1e-1 & 69.61\\ \hline\hline 
\end{tabular}}
\end{table}

We also experimented with adding an additional hidden layer for the Neural Network. Though, this did not improve on the MAPE scores.

\section{Results}

\autoref{tab:structured} shows results obtained from different models evaluated on the test set from online workload B. From the table, we can see that replacing K-means by GMM clustering improved the performance of the model to a slight extent by both metrics MAPE and MSE. However, the performance of the random forest algorithm is not as good as that of GPR. The neural network based model showed improved performance where in it was able to capture complex relationships which showed superior results in evaluations.This can be clearly seen with the MSE values seen in the \autoref{tab:structured}. However, it should be noted that neural networks have a higher tendency to overfit when given less data. Our experiments in general show that results are improved by using neural networks as well as GMM clustering.

\begin{table}[htbp]
\caption{Results summary}
\label{tab:structured}
\centering
\adjustbox{max width=\linewidth}{\fontsize{9}{12}\selectfont \it
\begin{tabular}{M{3.25cm}M{2.5cm}M{2.5cm}} \hline \hline
\textbf{Type} & \textbf{MAPE} & \textbf{MSE} \\ \hline
Baseline &  69.61 & 2118 \\ 
EM clustering & 67.85 & 2329 \\ 
Random Forest & 71.98 & 2846 \\  
Neural Network &  65.26 & 1965 \\ \hline\hline 
\end{tabular}}
\end{table}

\begin{figure}[htbp]
\centering
\begin{tabular}{c}
\includegraphics[width=0.9\linewidth,height=0.18\paperheight]{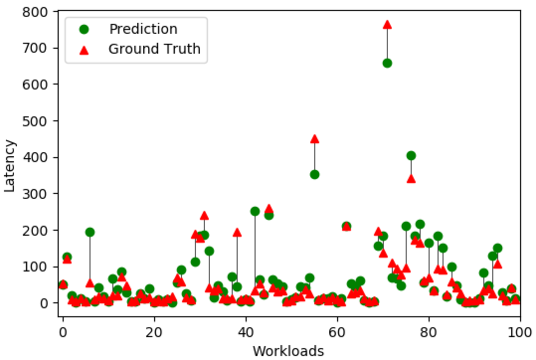} \\
(a) \\ [0.5mm]
\includegraphics[width=0.9\linewidth,height=0.18\paperheight]{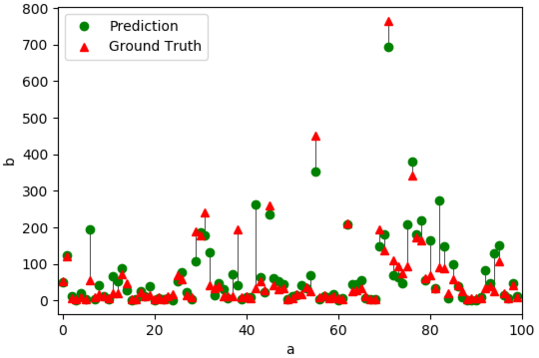} \\ 
(b) \\ [0.5mm]
\includegraphics[width=0.9\linewidth,height=0.18\paperheight]{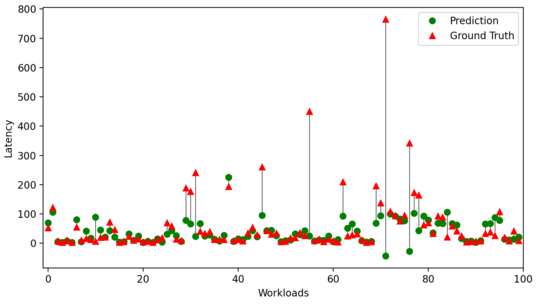} \\ 
(c) \\ [0.5mm]
\includegraphics[width=0.9\linewidth,height=0.18\paperheight]{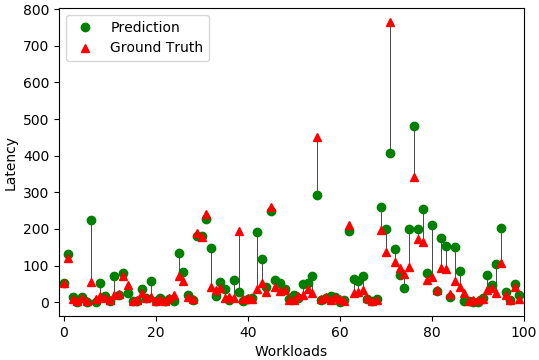} \\ 
(d) 
\end{tabular}
\caption{Predictions and ground truth latency prediction on a test set of online workload B (a) neural networks (b) EM clustering (c) random forest (d) baseline}
\label{fig:baseline_emclustering_NN-2_random}
\end{figure}

\autoref{fig:baseline_emclustering_NN-2_random} shows the ground truth and model predictions for latency on online workload B for both baseline, EM clustering, and neural networks. The plot shows that the predictions made using the metrics from EM-clustering algorithms were much closer to the ground truth when compared with the baseline approach. However, the predictions are very far apart in the case of neural networks.

\section{Conclusion}

Automatic DBMS tuning remains an active area of research, and we presented an automatic approach that leverages past experience and collects new information to tune DBMS configurations. During experiments, we were able to achieve a MAPE of 69\% using the baseline implementation, which utilizes FA,K-means for metric pruning, and GPR for prediction modeling.

Our experiments also show that by replacing K-means clustering with EM-clustering, we were able to reduce MAPE to 67\% which suggests that GMM clustering could be  an alternative to K-means clustering. The deep learning-based approach performed better than both of these clustering approaches, leading to a MAPE score of 65\%. Neural networks can perform better than traditional machine learning clustering models for the use case mentioned above because of their ability to handle high-dimensional data and identify complex patterns in it.


\begin{thebibliography}{00}

\bibitem{Dana_Van} Dana Van Aken, Andrew Pavlo, Geoffrey J. Gordon, and Bohan Zhang. 2017. \href{https://doi.org/10.1145/3035918.3064029}{Automatic database management system tuning through large-scale machine learning}. \textit{In Proceedings of the 2017 ACM International Conference on Management of Data}, SIG- MOD ’17, page 1009–1024, New York, NY, USA. Association for Computing Machinery.

\bibitem{Dworin} D. Dworin. { Data science revealed: A data-driven glimpse into the burgeoning new field.} Dec. 2011.

\bibitem{Laney} D. Laney.{ 3-D data management: Controlling data volume, velocity and variety.} Feb. 2001.

\bibitem{24_cite} S. Duan, V. Thummala, and S. Babu. { Tuning database configuration parameters with iTuned.}. \textit{VLDB, 2:1246–1257,} August 2009.

\bibitem{22_cite} K. Dias, M. Ramacher, U. Shaft, V. Venkataramani, and G. Wood. { Automatic performance diagnosis and tuning in oracle.}. \textit{CIdR}, 2005.

\bibitem{47_cite}  A. J. Storm, C. Garcia-Arellano, S. S. Lightstone, Y. Diao, and M. Surendra. { Adaptive self-tuning memory in DB2}. \textit{VLDB, pages 1081–1092} August 2006.

\bibitem{36_cite}  M. Linster. { Best practices for becoming an exceptional postgres dba}. \textit{VLDB, 2:1246–1257,} August 2014.\url{http://www.enterprisedb.com/best-practices-becoming-exceptional-postgres-dba}

\bibitem{1_cite} {MySQL – InnoDB startup options and system variables}. \url{http://dev.mysql.com/doc/refman/5.6/en/innodb-parameters.html}


\bibitem{paper_12} P. Bernstein, M. Brodie, S. Ceri, D. DeWitt, M. Franklin, H. Garcia-Molina, J. Gray, J. Held, J. Hellerstein,
H. Jagadish, et al. The asilomar report on database research. \textit{SIGMOD record}, 27(4):74–80, 1998.

\bibitem{paper_18} S. Chaudhuri and G. Weikum. Rethinking database system architecture: Towards a self-tuning RISC-style
database system. In \textit{VLDB}, pages 1–10, 2000.

\bibitem{paper_16} S. Chaudhuri and V. Narasayya. Self-tuning database systems: a decade of progress. In \textit{VLDB}, pages 3–14,
2007.

\bibitem{paper_74}   G. Weikum, A. Moenkeberg, C. Hasse, and P. Zabback. Self-tuning database technology and information
services: From wishful thinking to viable engineering. In \textit{Proceedings of the 28th International Conference
on Very Large Data Bases}, VLDB ’02, pages 20–31, 2002.

\bibitem{paper_14} S. Ceri, S. Navathe, and G. Wiederhold. Distribution design of logical database schemas. \textit{IEEE Trans. Softw.
Eng}., 9(4):487–504, 1983.

\bibitem{paper_24} S. Duan, V. Thummala, and S. Babu. Tuning database configuration parameters with iTuned. \textit{VLDB},
2:1246–1257, August 2009.

\bibitem{paper_82} D. C. Zilio, J. Rao, S. Lightstone, G. Lohman, A. Storm, C. Garcia-Arellano, and S. Fadden. DB2 design
advisor: integrated automatic physical database design. In \textit{VLDB}, pages 1087–1097, 2004.

\bibitem{paper_75} D. Wiese, G. Rabinovitch, M. Reichert, and S. Arenswald. Autonomic tuning expert: A framework for
best-practice oriented autonomic database tuning. In \textit{Proceedings of the 2008 Conference of the Center for
Advanced Studies on Collaborative Research: Meeting of Minds}, CASCON ’08, pages 3:27–3:41, 2008.

\bibitem{paper_32} M. Hammer and B. Niamir. A heuristic approach to attribute partitioning. In \textit{SIGMOD}, pages 93–101, 1979.

\bibitem{paper_42} E. Kwan, S. Lightstone, A. Storm, and L. Wu. Automatic configuration for IBM DB2 universal database.
Technical report, IBM, jan 2002.

\bibitem{paper_67} A. J. Storm, C. Garcia-Arellano, S. S. Lightstone, Y. Diao, and M. Surendra. Adaptive self-tuning memory in DB2. In \textit{VLDB}, pages 1081–1092, 2006

\bibitem{paper_70} W. Tian, P. Martin, and W. Powley. Techniques for automatically sizing multiple buffer pools in DB2. In \textit{CASCON}, pages 294–302, 2003.  

\bibitem{paper_19} B. Dageville and M. Zait. Sql memory management in oracle9i. \textit{In Proceedings of the 28th International Conference on Very Large Data Bases}, VLDB ’02, pages 962–973, 2002.

\bibitem{paper_21} K. Dias, M. Ramacher, U. Shaft, V. Venkataramani, and G. Wood. Automatic performance diagnosis and
tuning in oracle. In \textit{CIdR}, 2005.

\bibitem{paper_40} S. Kumar. Oracle Database 10g: The self-managing database, Nov. 2003. White Paper

\bibitem{paper_17}  S. Chaudhuri and V. R. Narasayya. An efficient cost-driven index selection tool for microsoft SQL server. In \textit{VLDB}, pages 146–155, 1997.

\bibitem{paper_58} A. Pavlo, C. Curino, and S. Zdonik. Skew-Aware Automatic Database Partitioning in Shared-Nothing,
Parallel OLTP Systems. In \textit{SIGMOD}, pages 61–72, 2012.

\bibitem{paper_54} R. Nehme and N. Bruno. Automated partitioning design in parallel database systems. In \textit{SIGMOD}, SIGMOD, pages 1137–1148, 2011.

\bibitem{paper_77} B. Xi, Z. Liu, M. Raghavachari, C. H. Xia, and L. Zhang. A smart hill-climbing algorithm for application
server configuration. In \textit{WWW}, pages 287–296, 2004.

\bibitem{paper_13} N. Bruno and S. Chaudhuri. Automatic physical database tuning: a relaxation-based approach. In \textit{SIGMOD}, pages 227–238, 2005.

\bibitem{paper_15} S. Chaudhuri and V. Narasayya. Autoadmin “what-if” index analysis utility. \textit{SIGMOD Rec}., 27(2):367–378, 1998.

\bibitem{paper_65}  A. A. Soror, U. F. Minhas, A. Aboulnaga, K. Salem, P. Kokosielis, and S. Kamath. Automatic virtual machine configuration for database workloads. In \textit{SIGMOD}, pages 953–966, 2008.

\bibitem{paper_53} D. Narayanan, E. Thereska, and A. Ailamaki. Continuous resource monitoring for self-predicting DBMS. In \textit{MASCOTS}, pages 239–248, 2005.

\bibitem{paper_2} MySQL Tuning Primer Script. \url{https://launchpad.net/mysql-tuning-primer}.

\bibitem{paper_7} PostgreSQL Configuration Wizard. \url{https://pgtune.leopard.in.ua}.

\bibitem{paper_52} R. Mukkamala, S. C. Bruell, and R. K. Shultz. Design of partially replicated distributed database systems: an integrated methodology. \textit{SIGMETRICS Perform. Eval. Rev}., 16(1):187–196, 1988.

\bibitem{paper_35} M. Y. L. Ip, L. V. Saxton, and V. V. Raghavan. On the selection of an optimal set of indexes. \textit{IEEE Trans. Softw. Eng}., 9(2):135–143, 1983.

\bibitem{sklearn_FA} Factor analysis implementation from sklearn. \url{https://scikit-learn.org/stable/modules/generated/sklearn.decomposition.FactorAnalysis.html}

\bibitem{sklearn_kmeans} K-means clustering implementation from sklearn. \url{https://scikit-learn.org/stable/modules/generated/sklearn.cluster.KMeans.html}

\bibitem{sklearn_GPR} Guassian process regression implementation from sklearn. \url{https://scikit-learn.org/stable/modules/gaussian_process.html}

\bibitem{sklearn_GMM} Guassian mixture model (em-clustering) implementation from sklearn. \url{https://scikit-learn.org/stable/modules/generated/sklearn.mixture.GaussianMixture.html#sklearn.mixture.GaussianMixture}

\bibitem{sklearn_RF} Random forest sklearn implementation. \url{https://scikit-learn.org/stable/modules/generated/sklearn.ensemble.RandomForestRegressor.html}

\end{thebibliography}
\end{document}